\PassOptionsToPackage{table}{xcolor}
\documentclass{bmvc2k}

\usepackage[table]{xcolor}
\usepackage{adjustbox}
\definecolor{red}{rgb}{0.95,0.4,0.4}
\definecolor{blue}{rgb}{0.4,0.4,0.95}
\definecolor{darkblue}{rgb}{0,0,0.8}
\definecolor{purple}{RGB}{195, 177, 225	}
\definecolor{darkgreen}{RGB}{119,221,119}
\definecolor{grey}{rgb}{0.6,0.6,0.6}
\definecolor{col1}{RGB}{232, 161, 148}
\definecolor{col2}{RGB}{167, 199, 231}
\definecolor{pink}{RGB}{248, 200, 220}
\usepackage{pifont}
\newcommand{\xmark}{\ding{55}}%
\newcommand{\cmark}{\ding{51}}%

\title{BaseBoostDepth: Exploiting Larger Baselines For Self-supervised Monocular Depth Estimation}

\addauthor{Kieran Saunders}{190229315@aston.ac.uk}{1}
\addauthor{Luis J. Manso}{l.manso@aston.ac.uk}{1}
\addauthor{George Vogiatzis}{g.vogiatzis@lboro.ac.uk}{2}

\addinstitution{
 Aston University\\
 Birmingham, UK
}
\addinstitution{
 Loughborough University\\
 Leicestershire, UK
}

\runninghead{Saunders, Manso, Vogiatzis}{BaseBoostDepth}

\def\etal{\emph{et al}\bmvaOneDot}
\begin{document}

\maketitle

\begin{abstract}
    In the domain of multi-baseline stereo, the conventional understanding is that, in general, increasing baseline separation substantially enhances the accuracy of depth estimation. However, prevailing self-supervised depth estimation architectures primarily use minimal frame separation and a constrained stereo baseline. Larger frame separations can be employed; however, we show this to result in diminished depth quality due to various factors, including significant changes in brightness, and increased areas of occlusion. In response to these challenges, our proposed method, \textbf{BaseBoostDepth}, incorporates a curriculum learning-inspired optimization strategy to effectively leverage larger frame separations. However, we show that our curriculum learning-inspired strategy alone does not suffice, as larger baselines still cause pose estimation drifts. Therefore, we introduce incremental pose estimation to enhance the accuracy of pose estimations, resulting in significant improvements across all depth metrics. Additionally, to improve the robustness of the model, we introduce error-induced reconstructions, which optimize reconstructions with added error to the pose estimations. Ultimately, our final depth network achieves state-of-the-art performance on KITTI and SYNS-patches datasets across image-based, edge-based, and point cloud-based metrics without increasing computational complexity at test time. The project website can be found at \href{https://kieran514.github.io/BaseBoostDepth-Project/}{https://kieran514.github.io/BaseBoostDepth-Project/}. 
\end{abstract}

\section{Introduction}
\label{sec:intro}
\begin{figure}[ht]
  \centering
   \includegraphics[width=1\linewidth]{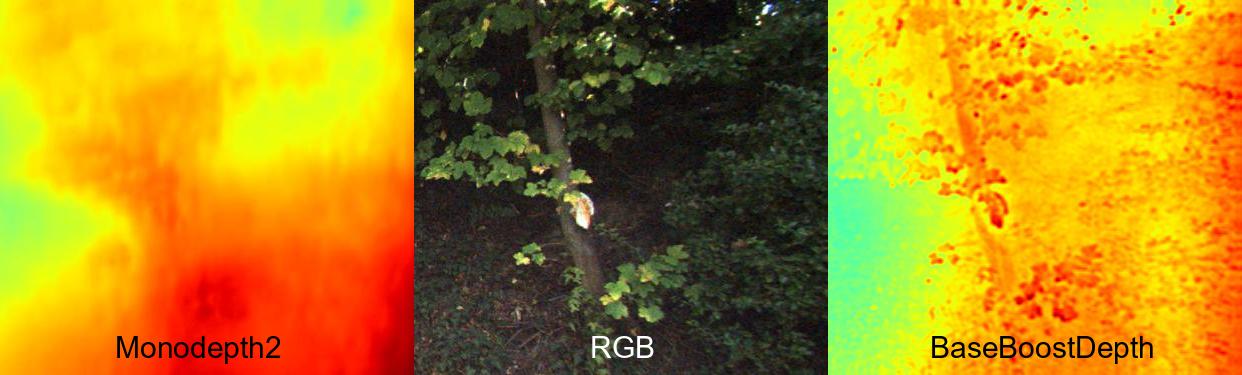}
   \caption{When comparing \textbf{BaseBoostDepth} with the baseline Monodepth2, we observe significant improvements in edge-based depth estimation metrics.}
   \label{fig:onecol}
    \vspace{-0.4cm}
\end{figure}

For decades, depth estimation has stood as a fundamental element in the domain of computer vision, finding diverse applications in areas like self-driving, virtual reality, robotics, and scene reconstruction. While the principles of multiple view geometry have long been understood, the rise of deep learning has made single-view depth prediction feasible. 

Most self-supervised approaches to monocular depth estimation use photometric loss to evaluate view-synthesis between consecutive video frames, deviating from traditional supervised learning that relies on significant ground truth depth data obtained from expensive sensors like LiDAR. Self-supervised methods have attracted attention for their cost efficiency, as they remove the need for ground truth data. Consequently, they can be trained on larger datasets owing to the abundance of available video data, leading to enhanced generalizability as shown in prior research \cite{hu2020seasondepth}, compared to their supervised counterparts.

However, the significance of baseline width in self-supervised methods has not been explored to the same extent as it has in the field of multi-baseline stereo. In multi-baseline stereo, a consistent trend is known: narrower baselines pose an easier pixel matching problem but result in poorer depth estimates.

Despite the potential accuracy advantages of wider baselines, current self-supervised monocular depth (SSMD) methods, such as Monodepth2 (MD2) \cite{godard2019digging}, use narrower baselines in their reconstruction processes. MD2 does this using source images which consist of one subsequent and one preceding consecutive frame to reconstruct the target image. Additionally, it leverages narrow stereo frames in relation to the target image to aid in the reconstruction process. While it is possible to use larger monocular baselines, research conducted by Lokender \etal \cite{tiwari2020pseudo} has suggested that employing wider baselines over a larger temporal window introduces challenges such as brightness inconsistencies and increased occlusions, thus making the use of larger baselines a complex problem. 

One might consider a straightforward approach: combining large and small baselines and updating the depth estimation based on the most accurate image reconstruction. However, as demonstrated in Section \ref{ablation_study}, this approach introduces a significant bias in favor of smaller baselines, as depth inaccuracies in those images yield lower photometric errors.

Brightness-contrast cues \cite{schwartz1983luminance, tai2012luminance, hibbard2023luminance}, which play a crucial role in our method, rely on the fact that objects closer to the camera tend to appear brighter than those farther away. Additionally, while traditional image-based metrics have proven useful, we aim to bolster the case for wider baselines by also examining edge-based metrics \cite{koch2018evaluation}, providing a more accurate depiction of how humans perceive depth from two-dimensional images. Furthermore, we analyze point cloud metrics \cite{ornek20222d} to validate the suitability of our depth estimations for use in 3D applications.

In this work, we leverage wide monocular baselines to achieve state-of-the-art (SotA) depth predictions, as depicted in Figure \ref{fig:onecol}. Our proposed method, \textbf{BaseBoostDepth}, outperforms MD2 in terms of image and edge-based metrics. Distinctively, our approach exhibits a stronger reliance on brightness-contrast cues extracted from the input image. These cues significantly enhance boundary definition in our depth estimations without any edge-based supervision. To our knowledge, we are the first to observe the significance of brightness-contrast cues in SSMD estimation.

To accomplish this, we put forward four main contributions: 
\begin{itemize}
    \item \textbf{Curriculum-Learning-Inspired Optimization Strategy (\ref{rand})} – This strategy involves a gradual transition from smaller to wider monocular baselines through two stages of training: warmup and boosting.
    \item \textbf{Tri-Minimization (\ref{tri-min_method})} – Inspired by multi-baseline stereo, we minimize errors by reconstructing the target image (center frame) from triplets of future and past frames, effectively using multiple reconstructions from different baselines.
    \item \textbf{Incremental Pose Estimations (\ref{pose})} – To address significant drift in pose estimation over larger baselines, which tends towards underestimation, we introduce incremental pose estimation. This technique involves breaking down the pose estimation process into smaller increments within larger intervals.
    \item \textbf{Error-Induced Reconstructions (\ref{Error})} – In addition to using incremental pose estimation, we optimize reconstructions by applying controlled error to the pose estimates. This approach is motivated by our observation that incremental pose estimation does not benefit all reconstructions.
\end{itemize}
To systematically evaluate each contribution, we conduct an ablation study in Section \ref{ablation_study} and show SotA performance on both the KITTI and SYNS datasets.
\section{Related Work}
\textbf{Self-supervised Depth:}
Garg \etal \cite{garg2016unsupervised} pioneered self-supervised learning for stereo depth via view synthesis between stereo pairs. Subsequently, Monodepth \cite{godard2017unsupervised} used photometric loss, combining $L_1$ loss and SSIM \cite{wang2004image}, to enforce left-right consistency in reconstructed images. Our focus is on monocular cameras due to their inherent simplicity, which contrasts with the cost constraints and spatial limitations associated with stereo setups. SfM-Learner~\cite{zhou2017unsupervised} was the first to utilize view synthesis for monocular depth estimation. Unlike traditional stereo methods, this and subsequent approaches leverage a depth network along with pose estimations to warp images, thereby maximizing photometric uniformity. MD2 \cite{godard2019digging} introduced per-pixel minimization of photometric error to address occlusion issues, incorporating auto-masking for textureless regions, stationary pixels and dynamic objects.
Improving upon MD2, some methods have proposed better depth network architectures \cite{zhao2022monovit, wang2023sqldepth, zhou2021self, yan2021channel, zhang2023lite}, or introduced cost volumes to utilize multiple frames as input \cite{watson2021temporal, wimbauer2021monorec, miao2023ds}. Other methods have focused on improving the robustness of monocular depth estimation \cite{zheng2024physical, saunders2023self, wang2024digging, mao2024stealing, zhu2023ec}, or on handling the rigid scene assumption \cite{lee2021learning, saunders2023dyna, feng2022disentangling, casser2019unsupervised}.

\textbf{Wider Baselines \& Brightness-contrast Cues:}
Lokender \etal \cite{tiwari2020pseudo} were among the first to propose the use of larger frame separations to enhance depth accuracy. However, concerns were raised that larger frame separations would introduce challenges due to increased occlusion and brightness inconsistencies. Madhu \etal \cite{vankadari2023sun} tackled brightness inconsistencies by implementing per-pixel neural intensity transformation, allowing for a two-frame separation instead of one. Their findings suggested that this increased separation led to improved depth, particularly when addressing brightness inconsistencies. Notably, unlike our method, their exploration did not extend to larger frame separations.
\section{Method:}
\begin{figure*}[th!]
  \centering
    \includegraphics[width=1\linewidth]{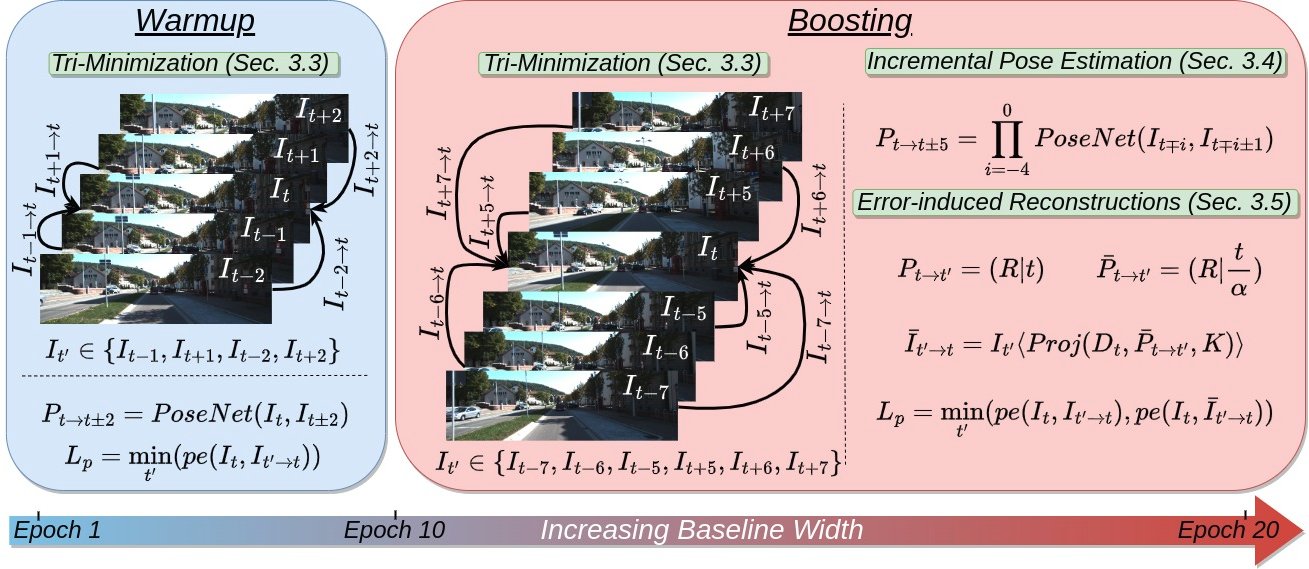}
       \caption{We progressively increase the baseline width used for training the depth and pose networks (\ref{rand}). We employ tri-minimization in both warmup and boosting stages (\ref{tri-min_method}), incorporating incremental pose estimation (\ref{pose}) and error-induced reconstructions (\ref{Error}) exclusively during the boosting stage.}
   \label{fig:method}
\end{figure*}

\textbf{Overview:}
We present \textbf{BaseBoostDepth}, which uses a curriculum-learning-inspired optimization strategy divided into warm-up and boosting stages. Our approach is capable of accurately estimating depth with clearly defined object boundaries. Unlike previous methods, we effectively exploit wider baselines and observe a greater effect of brightness-contrast cues, resulting in SotA depth estimations. Our method is depth backbone-agnostic, allowing any pre-trained or from-scratch depth network to be boosted and achieve enhanced object boundary definition. An overview of the overall framework is depicted in Figure \ref{fig:method}.
\subsection{Preliminaries}
Adhering to the methodology outlined by Zhou \etal \cite{zhou2017unsupervised}, we concurrently train both an ego-motion network and a depth network to facilitate view synthesis between successive frames. Our approach entails using the target depth estimation $D_{t}=DepthNet(I_t)$ and camera pose estimations $P_{t \to t^\prime} = PoseNet(I_t,I_{t^\prime})$ to synthesize the target image, where $I_{t^\prime} \in \{I_{t-1},I_{t+1}\}$, relying solely on source frames. Here, PoseNet and DepthNet denote the pose and depth estimation networks, respectively. The synthesized projection is obtained through inverse warping, as illustrated in the equation below.
\begin{equation} \label{eq1}
I_{{t^\prime} \to t} = I_{t^\prime} \langle Proj(D_{t}, P_{t \to {t^\prime}}, K) \rangle
\end{equation}
The Proj() function yields the resulting 2D coordinates of the depths after projecting into the camera of frame $I_{t^\prime}$, with $\langle \rangle$ denoting the sampling operator. Furthermore, following from \cite{godard2017unsupervised}, our methodology incorporates photometric loss ($pe$), which is defined as follows:
\begin{equation} \label{pe}
pe(I_a,I_b)=\frac{0.85}{2}\Big(1-SSIM(I_a,I_b)\Big)+(1-0.85)||I_a-I_b||.
\end{equation}
In MD2, the per-pixel photometric loss employed for training the pose and depth network uses minimum aggregation, as depicted below:
\begin{equation} \label{min}
L_p = \min_{t^\prime}(pe(I_t, I_{t^\prime \to t})).
\end{equation}
For each pixel, a determination is made regarding whether to use the next or the previous frames for reprojection based on the minimization of reprojection errors. 
\subsection{Curriculum-learning-inspired Optimization Strategy}\label{rand}
Inspired by curriculum learning's gradual progression from easier to harder samples during training, we categorize image reconstructions by baseline widths. "Easy" samples have smaller baselines, while "harder" samples have larger baselines. Increasing frame separation can yield larger baselines, but baseline widths vary significantly. For example, with a two-frame separation, baselines range from approximately 0.1 to 0.8 distance units, with 0.1 representing roughly 0.54m (stereo images baseline, $I_s$).

We select frame separations for reconstruction based on predetermined distance estimation using a pre-trained pose network (MD2) between each frame and its subsequent frame in the training dataset. The $L_2$ norm of the estimated translation matrix determines the baseline for each frame separation, denoted as $b$ for one frame of separation. To calculate the baseline between the target and potential source images, we multiply the predetermined baseline ($b$) by the number of frames ($k$) separating them. This can be represented as follows:
\begin{equation}
G(t,t+k) = b \times k.
\end{equation}
Where $G(t,s) = 0.1$ for stereo frames. This assumes approximately constant velocity over small frame separations.

We select the source frames index denoted by $\hat{x}^+$ to reconstruct the target image $I_t$, adhering to the following equation:
\begin{align}\label{warmupeq}
    \hat{x}^+ = \arg\max_{x} \left( G(t, x) \ \middle|\ x \in \Omega, G(t, x) <= \tau \right) 
\end{align}
Here, we choose the source image index ($\hat{x}^+$) with the highest baseline relative to the target image ($G(t,x)$), provided it falls below the threshold ($\tau$), from a predefined set of potential source images ($\Omega$). We set $\tau$ and $\Omega$ to $0.1 + (0.04 \times \text{epoch})$ and $ \{s,{t+1},{t+2}\}$ respectively for the warm-up stage and to $(0.1 \times \text{epoch}) - 0.4 $ and $ \{s,{t+1},{t+2},{t+3},{t+4},{t+5}\}$ respectively for the boosting stage. To mitigate external influences like lighting variations and dynamic objects, we limit our model to a maximum of two frames during the warm-up stage.

We use the positive frame indices as shown in Eq. \ref{warmupeq}, but we also include the corresponding negative versions of the monocular images. This leads to an updated set of source images $I_{t^\prime} \in \{ I_{\hat{x}^+}, I_{\hat{x}^-} \}$ when a monocular frame index is chosen, where $\hat{x}^+ = t+k$ and $\hat{x}^- = t-k$. If a stereo index is selected as the source frame, then $I_{t^\prime} \in \{ I_{\hat{x}^+} \}$.


In summary, for each target frame in a batch, we select the maximum frame separation within the bounds of Eq. \ref{warmupeq}, which leads to a varied set of source images $I_{t^\prime}$ across the batch. 
\subsection{Tri-minimization:}\label{tri-min_method}
We introduce tri-minimization, a technique to reconstruct the target image (center frame) from triplets of future and past frames with different baselines to address occlusion, mitigate brightness inconsistencies, and reduce the impact of dynamic objects. 

To achieve tri-minimization we attempt to extend the set of source image to include three future frames and three previous frames, including the selected source frame from Eq. \ref{warmupeq}. The formal equation is shown below:
\begin{align}\label{optim_1}
    I_{t^\prime} \in \begin{cases}
    \{I_{\hat{x}^+}\},      & \text{if } \hat{x}^+ = s \\
    \{I_{\hat{x}^+}, I_{\hat{x}^-}, I_s\},      & \text{if } \hat{x}^+ = {t+1} \\
    \{I_{\hat{x}^+},I_{\hat{x}^+-1}, I_{\hat{x}^-}, I_{\hat{x}^-+1}, I_s\},   & \text{if } \hat{x}^+ = {t+2} \\
    \{I_{\hat{x}^+},I_{\hat{x}^+-1},I_{\hat{x}^+-2}, I_{\hat{x}^-}, I_{\hat{x}^-+1},I_{\hat{x}^-+2}\},  & \text{otherwise. }
\end{cases}
\end{align}
Given that a monocular frame is selected (i.e., $\hat{x}^+ \neq s$), then $\hat{x}^-+1 = t-k+1$ and $\hat{x}^+-1 = t+k-1$. When using tri-minimization, we encourage larger baseline widths to counteract the preference for smaller baselines in minimization aggregation. This is achieved by using more widely separated potential source frames ($\Omega$) and a more aggressive $\tau$ threshold in the boosting stage ($\Omega = \{s,{t+1},{t+2},{t+3},{t+4},{t+5},{t+6},{t+7}\}$ and $\tau = (0.15 \times \text{epoch}) - 0.9$).
\subsection{Incremental Pose Estimation}\label{pose}
When training with larger frame separations and assuming constant velocity within short time intervals (less than one second), consecutive frame translations are expected to remain approximately linear. However, upon inspection, we observed pose estimation drift with increased frame separations, indicating better performance with smaller separations and worse with larger ones. Therefore, we propose incremental pose estimation as follows:
\begin{align}\label{coirrect}
    {P}_{t \to t \pm n} = \prod_{i=-(n-1)}^{0} PoseNet(I_{t \mp i}, I_{t \mp i \pm 1}).
\end{align}
Equation \ref{coirrect} represents the matrix multiplication of incremental pose estimations, leading to a refined pose estimation over larger frame separations. For further support, see the supplementary materials. 

During tri-minimization, we discovered that using incremental pose estimations was beneficial only for the smallest frame separation to the target image within the set of source images $I_{t^\prime}$. However, we found that rotation estimations from incremental pose estimations were beneficial for all reconstructions.
\subsection{Error-induced Reconstructions}\label{Error}
Based on the discovery that partial incremental pose results in better image-based and edge-based performance than a full incremental pose, we propose that adding a fixed error to the pose network could lead to improved performance. Integrating reconstructions based on pose estimations with a fixed error in translations empirically enables the depth network to better understand the influence of pose estimations on reconstruction accuracy. By incorporating these reconstructions, we expand the solution space, where the introduced pose errors act as a form of perturbation to guide the depth network towards exploring alternative solutions and enhance its ability to generalize. This observation is well-supported by experimentation. Refer to Section \ref{ablation_study} for more details.

Using the incremental pose estimations, we define the rotation and translation as $(R|t) = P_{t \to {t^\prime}}$, then the error-induced pose is defined as $\bar{P}_{t \to {t^\prime}} = (R|\frac{t}{\alpha})$. Then;
\begin{align}
    \bar{I}_{{t^\prime} \to t} = I_{t^\prime} \langle Proj(D_{t}, \bar{P}_{t \to {t^\prime}}, K) \rangle.
\end{align}
Note that we do not change the corrected rotation estimations and that the error-induced pose gradients are cut off during backpropagation. Finally, we minimize between the standard reconstructions $pe(I_t, I_{t^\prime \to t})$ which use incremental pose and the error-induced reconstructions $pe(I_t, \bar{I}_{t^\prime \to t})$ which use the error-induced pose:
\begin{equation}\label{optim}
    L_p = \min_{t^\prime}(pe(I_t, I_{t^\prime \to t}), pe(I_t, \bar{I}_{t^\prime \to t})).
\end{equation}
The final loss, incorporating photometric loss with automasking ($\mu$) from MD2 \cite{godard2019digging} and per-pixel smoothness loss from \cite{ranjan2019competitive}, is defined as $L = \mu L_{p} + \lambda L_s$, and this combined loss is averaged across each pixel, scale, and batch.
\section{Results:}
\textbf{Experimental set-up:}
For training \textbf{BaseBoostDepth}, we utilize pretrained ImageNet weights \cite{deng2009imagenet} with PyTorch \cite{NEURIPS2019_9015} on an NVIDIA A6000 GPU. We employ the Adam optimizer \cite{kingma2014adam} for 20 epochs, using an input size of $640\times 192$ and a multi-step learning rate strategy. The learning rate starts from 1e-4 and is progressively reduced at epochs 11, 13, 15, 16, 17, 18, and 19 by a factor of 0.4. Hyperparameters $\omega$, $\beta$, and $\gamma$ are set to 0.01, 0.01, and 0.001, respectively, with a smoothing loss parameter $\lambda$ of 0.001. Through empirical testing, the warm-up stage spans the initial 10 epochs using 4 resolution scales, while the boosting stage covers the subsequent 10 epochs with only the largest resolution scale. For variations like \textbf{BaseBoostDepth$_{pre}$} and \textbf{BaseBoostDepth$_{pre}^\dagger$} in Table \ref{tab:kitti_eigen} and Table \ref{tab:SYNS}, we train exclusively with the boosting stage starting from pretrained weights and depth backbones from MD2 (ResNet-18 encoder) and MonoViT, respectively.
\subsection{Datasets}
\textbf{KITTI} \cite{geiger2012we}:
For validation, we use the official Zhou split with 4,424 images and train on the full set of 39,810 images. Testing is done on 697 images from the Eigen \etal test set \cite{eigen2015predicting}. Due to ground truth accuracy limitations, our evaluation focuses on image-based metrics for depth estimate assessment. All models are trained exclusively on the KITTI dataset.

\noindent
\textbf{SYNS-Patches} \cite{adams2016southampton}:
This dataset comprises 1,438 outdoor images with accurately measured ground truth depth information. We adopt edge-based metrics from Koch \etal \cite{koch2018evaluation} and point cloud-based metrics from Örnek \etal \cite{ornek20222d}, following the methodology outlined by Spencer \etal \cite{spencer2022deconstructing}. Note that the exact steps for evaluating this dataset are not provided; therefore, we have created our own version of the evaluation, which is released with the code.
\subsection{Ablation Study:}\label{ablation_study}
This subsection investigates the impact of each contribution on the baseline model, as shown in Table \ref{ablation}. We primarily analyze the KITTI test dataset using image-based metrics and evaluate edge-based metrics for each contribution using the SYNS test set. 
\begin{table*}[ht]
    \centering
    \begin{adjustbox}{max width=\textwidth}
    \large
    \begin{tabular}{|l|c|c|c|c|c|c||c|c|c|c|c|c|c||c|c|}\hline
    
       & \multicolumn{6}{c||}{Contributions} & \multicolumn{7}{c||}{KITTI} & \multicolumn{2}{c|}{SYNS} \\

        \textbf{Ablation} & 
        \cellcolor{purple}\begin{tabular}{@{}c@{}}Skip \end{tabular} & 
        \cellcolor{purple}\begin{tabular}{@{}c@{}}Pre \end{tabular} & 
        \cellcolor{purple}\begin{tabular}{@{}c@{}}Tri. \end{tabular} & 
        \cellcolor{purple}\begin{tabular}{@{}c@{}}Incri. \\ Pose \end{tabular} & 
        \cellcolor{purple}\begin{tabular}{@{}c@{}}Part. \\ Incri.\end{tabular} & 
        \cellcolor{purple}\begin{tabular}{@{}c@{}}Err. \\ Rec. \end{tabular} & 
         \cellcolor{col1}Abs Rel & 
         \cellcolor{col1}Sq Rel & 
         \cellcolor{col1}RMSE & 
        \cellcolor{col1}RMSE log & 
        \cellcolor{col2}$\delta < 1.25 $ & 
        \cellcolor{col2}$\delta < 1.25^{2}$ & 
        \cellcolor{col2}$\delta < 1.25^{3}$ & 
        \cellcolor{col1}Acc & 
        \cellcolor{col1}Comp  \\ 
        
        \hline
        
        Monodepth2 \cite{godard2019digging} & 1 & & & & & & 
         \underline{0.106} & 0.818 & 4.750 & 0.196 & 0.874 & 0.957 & 0.979 &   2.516  &  17.193 \\ \hline 
         
        Monodepth2 \cite{godard2019digging} & 4 & & & &  & &   {0.107}  &   0.832  &   4.723  &   {0.186}  &   \underline{0.887}  &   0.961  &   \underline{0.982} &   2.512  &  14.856\\ \hline 
        
        Monodepth2 \cite{godard2019digging} & [4] & & & &  & &     0.146  &   1.164  &   5.289  &   0.221  &   0.813  &   0.940  &   0.975 &   2.465  &   5.278 \\ \hline \hline
          
        BaseBoostDepth & $C$& \xmark  & & & & & 0.115  &   0.916  &   4.856  &   0.190  &   0.877  &   0.960  &   \textbf{0.983} &   {2.442}  &   5.518    \\ \hline 
        
        BaseBoostDepth  & $C$& \xmark  & \cmark & & &   & 0.112  &   0.867  &   4.762  &   0.187  &   0.879  &   \underline{0.962}  &   \textbf{0.983} &   \textbf{2.417}  &   \textbf{3.433}  \\ \hline
        
        BaseBoostDepth  & $C$& \xmark  & \cmark & \cmark &&&    0.109  &   0.868  &   4.767  &   {0.186}  &   0.883  &   0.961  &   \underline{0.982} &   2.489  &   6.547   \\ \hline
                
        BaseBoostDepth  & $C$& \xmark  & \cmark& \cmark & \cmark &   
        &   {0.107}  &   {0.799}  &   {4.656}  &   \underline{0.184}  &   {0.884}  &   \textbf{0.963}  &   \textbf{0.983} &   2.450  &   4.290   \\ \hline
        
        \textbf{BaseBoostDepth}  & $C$& \xmark  & \cmark& \cmark & \cmark &   \cmark  
        &   \underline{0.106}  &   \underline{0.736}  &   \underline{4.584}  &   \underline{0.184}  &   0.883  &   \textbf{0.963}  &   \textbf{0.983} &   2.453  &   \underline{3.810}  \\ \hline \hline
        
        \textbf{BaseBoostDepth$_{pre}$}  & $C$& \cmark  & \cmark& \cmark & \cmark &   \cmark  
        &   \textbf{0.104}  &   \textbf{0.738}  &   \textbf{4.544}  &   \textbf{0.183}  &   \textbf{0.888}  &   \textbf{0.963}  &   \textbf{0.983} & \underline{2.432}  &   {4.763}  \\ \hline

    \end{tabular}
    \end{adjustbox}
    \caption{\textbf{Ablation Study:} Here, we present all contributions of our work. \textbf{Bold} represents the best results for the metric and \underline{underscore} is the second best. }
    \label{ablation}
\end{table*}

\noindent
\textbf{Baseline (Row 1 \& Row 2):}
The first row depicts MD2 using a one-frame separation, trained with monocular and stereo images. In contrast, the second row involves reconstructions using up to 4-frames of separation, but shows no significant quantitative improvements in image-based or edge-based depth metrics. This highlights the challenge of using wider baselines with standard minimum aggregation, where smaller baselines produce more accurate reconstructions despite potentially less accurate depth estimations. As a result, wider baselines are often overlooked in optimization steps, a challenge addressed by \textbf{BaseBoostDepth}.

\noindent
\textbf{Wide Baseline (Row 3):}
Results are shown after training with a 4-frame separation ($I_{t^\prime} \in {I_{t-4},I_{t+4}}$). The network faces challenges with larger baselines due to increased occlusion, brightness changes, and dynamic objects. While image-based metrics decline, edge-based metrics improve significantly, indicating better edge definition with larger baselines.

\noindent
\textbf{Curriculum-learning-inspired Optimization (Row 4):}
Our curriculum-learning-inspired optimization strategy ($C$) yields worse image-based metrics compared to row 1 due to larger baselines introduced during the boosting phase, leading to unstable optimization. However, we maintain respectable image-based metrics and achieve significant edge-based improvements similar to row 3, thanks to the gradual introduction of larger baselines.

\noindent
\textbf{Tri-minimization (Row 5):}
Tri-minimization yields improvements in image-based metrics and notably achieves impressive edge-based metrics surpassing those of row 3. This shows our ability to achieve greater edge-based metrics while improving image-based metrics.

\noindent
\textbf{Incremental Pose Estimation (Row 6 \& Row 7):}
We hypothesize that many observed errors were due to drifted pose estimations. To tackle this, in row 6 we introduce incremental pose estimation, resulting in another decline in image-based metrics but an increase in edge-based metrics.
In row 7, we apply incremental pose solely to the smallest frame separation in tri-minimization, while consistently using incremental pose estimations for rotation (referred to as "partial incremental pose"). This approach again leads to significant reductions in both image-based and edge-based metrics.

\noindent
\textbf{Error-induced Reconstructions (Row 8):}
We observe another notable decrease in both edge-based and image-based metrics when using error-induced reconstructions with an $\alpha$ value of 5.5. Fine-tuning details are provided in the supplementary materials.

\noindent
\textbf{Pre-trained with MD2 (Row 9):}
Finally, we initialize our model with Monodepth2 weights and then apply our boosting stage. Combining contributions, we exceed the image-based metrics set by Monodepth2 and achieve significant improvements in edge-based performance.
\subsection{Comparison with SotA}
In Table \ref{tab:kitti_eigen}, we compare the previous SotA results with different variations of \textbf{BaseBoostDepth}. The version trained from scratch achieved performance comparable to MD2. However, significant performance gains were observed when applying the boosting phase to pre-trained depth networks. Our method consistently outperforms the original approach by leveraging all contributions from the boosting phase, and \textbf{BaseBoostDepth$_{pre}^\dagger$} establishes a new SotA benchmark for the given resolution using the MonoViT depth backbone and pre-trained weights.
\begin{table*}[ht]
  \centering
  \scriptsize
  \resizebox{1.0\textwidth}{!}{
    \begin{tabular}{|l||c|c|c|c|c|c|c|}
        \arrayrulecolor{black}\hline
          Method & \cellcolor{col1}Abs Rel & \cellcolor{col1}Sq Rel & \cellcolor{col1}RMSE  & \cellcolor{col1}RMSE log & \cellcolor{col2}$\delta < 1.25 $ & \cellcolor{col2}$\delta < 1.25^{2}$ & \cellcolor{col2}$\delta < 1.25^{3}$ \\
         
        \hline\hline
        
        Monodepth2~\cite{godard2019digging}  &
         0.106 & 0.818 & 4.750 & 0.196 & 0.874 & 0.957 & 0.979 \\ 
        \hline

        CADepth~\cite{yan2021channel}  &  0.102 & 0.752 &4.504 &0.181 &0.894 &0.964 &0.983 \\
        \hline
              
        DIFFNet ~\cite{zhou2021self} & {0.101} & 0.749 & 4.445 & 0.179 & {0.898} & 0.965 & 0.983 \\
        \hline
                 
        MonoViT~\cite{zhao2022monovit}  & \underline{0.098} & \underline{0.683} & \underline{4.333} & \underline{0.174} &\underline{0.904} &\underline{0.967} &\underline{0.984} \\
        
        \hline\hline
                
        \textbf{BaseBoostDepth}     &   0.106  &   {0.736}  &   4.584  &   0.184  &   0.883  &   {0.963}  &   {0.983}  \\
        \hline
        
        \textbf{BaseBoostDepth$_{pre}$}     &   0.104  &   0.738  &   4.544  &   0.183  &   0.888  &   0.963  &   0.983  \\
        \hline
        
        \textbf{BaseBoostDepth$_{pre}^\dagger$}     &   \textbf{0.096}  &   \textbf{0.648}  &   \textbf{4.201}  &   \textbf{0.170}  &   \textbf{0.906}  &   \textbf{0.968}  &   \textbf{0.985}  \\
       \hline
 
                
        \arrayrulecolor{black}\hline

    \end{tabular}}  
  \caption{\textbf{Quantitative Results for the KITTI Eigen Test Dataset.} Note that we do not use any test-time refinement processes. Our depth estimation inference relies on a single frame, and we train at a resolution of 640 × 192 without edge supervision. Models subscript with $pre$ are trained only with the boosting stage from pretrained weights, while $\dagger$ indicates the use of MonoViT's depth backbone; otherwise, a ResNet18 backbone is used.}
  \label{tab:kitti_eigen}
\end{table*}
\subsection{Evaluation of Edge and Point Cloud Performance}
\begin{table*}[ht]
  \centering
  \small
  \resizebox{1.0\textwidth}{!}{
    \begin{tabular}{|l||c|c|c|c|c||c|c||c|c|}
        \arrayrulecolor{black}\hline
          & \multicolumn{5}{c||}{Image-Based } & \multicolumn{2}{c||}{Edge-Based} & \multicolumn{2}{c|}{Point Cloud-Based} \\

         Method & \cellcolor{col1}Abs Rel&\cellcolor{col1}MAE&\cellcolor{col1}Sq Rel & \cellcolor{col1}RMSE & \cellcolor{col1}RMSE log  & \cellcolor{col1}Acc & \cellcolor{col1}Comp  & \cellcolor{col2}F-Score & \cellcolor{col2}IoU \\
         
        \hline\hline
        
        Monodepth2~\cite{godard2019digging}  
          &   0.334  &   6.901  &   5.285  &  12.089  &   {0.405}  &   2.516  &  17.193  &   0.242  &   0.149  \\ 
        \hline

        CADepth~\cite{yan2021channel}   &   0.363  &   8.787  &   5.548  &  13.512  &   0.546  &   2.473  &  19.045  &   0.022  &   0.012  \\

        \hline
        
        DIFFNet~\cite{zhou2021self}   &   {0.311}  &   6.554  &   4.690  &  11.610  &   0.383  &   \underline{2.411}  &  12.116  &   0.258  &   0.161  \\
        \hline
                               
        MonoViT~\cite{zhao2022monovit}   &   \underline{0.287}  &   \underline{6.195}  &   \underline{4.399}  &  \underline{11.124}  &   \underline{0.354}  &   2.443  &  15.672  &   0.264  &   0.164  \\
        
        \hline\hline
        
         \textbf{BaseBoostDepth}  &   0.334  &   6.878  &   4.854  &  11.847  &   {0.409}  &   {2.453}  &   \textbf{3.810}    &   \underline{0.275}  &   \underline{0.174}  \\ \hline
         
         \textbf{BaseBoostDepth$_{pre}$}  &   0.328  &   6.752  &   4.815  &  11.752  &   0.405  &   2.432  &   \underline{4.763}    &   0.268  &   0.168  \\ \hline

        \textbf{BaseBoostDepth$_{pre}^\dagger$}     &   \textbf{0.278}  &   \textbf{5.951}  &   \textbf{3.795}  &  \textbf{10.575}  &   \textbf{0.351}  &   \textbf{2.409}  &   5.314   &      \textbf{0.300}  &   \textbf{0.191} \\
        
        \hline 
        
        
        \arrayrulecolor{black}\hline

    \end{tabular}}  
      \caption{\textbf{Quantitative Results for the SYNS Test Dataset.} Here we show test evaluations done on the SYNS dataset using image, edge and point cloud  based metrics. 
      }
  \label{tab:SYNS} 
\end{table*}
To truly show the benefits of using large baselines, we evaluate our depth estimates against the SYNS-patches dataset to compare with other SotA models in Table \ref{tab:SYNS}.

Our analysis shows that leveraging our boosting phase improves performance across image, edge, and point cloud-based metrics on the SYNS dataset, \textit{regardless of the depth backbone used}. Comparing Monodepth2 with \textbf{BaseBoostDepth} trained using pre-trained ImageNet weights, we maintain similar image-based metric performance but achieve significant improvements in edge-based metrics and excel in point cloud metrics, demonstrating accuracy in 3D space.

In summary, \textbf{BaseBoostDepth$_{pre}^\dagger$} leads on the SYNS dataset, slightly behind in edge composition compared to \textbf{BaseBoostDepth}. 
Additionally, CNN-based architectures generally outperform vision transformers in edge composition, likely due to the texture-focused nature of CNNs versus the shape-focused approach of transformers. Ultimately, the choice of depth backbone depends on specific user requirements.

\begin{figure*}[th!]
  \centering
    \includegraphics[width=1\linewidth]{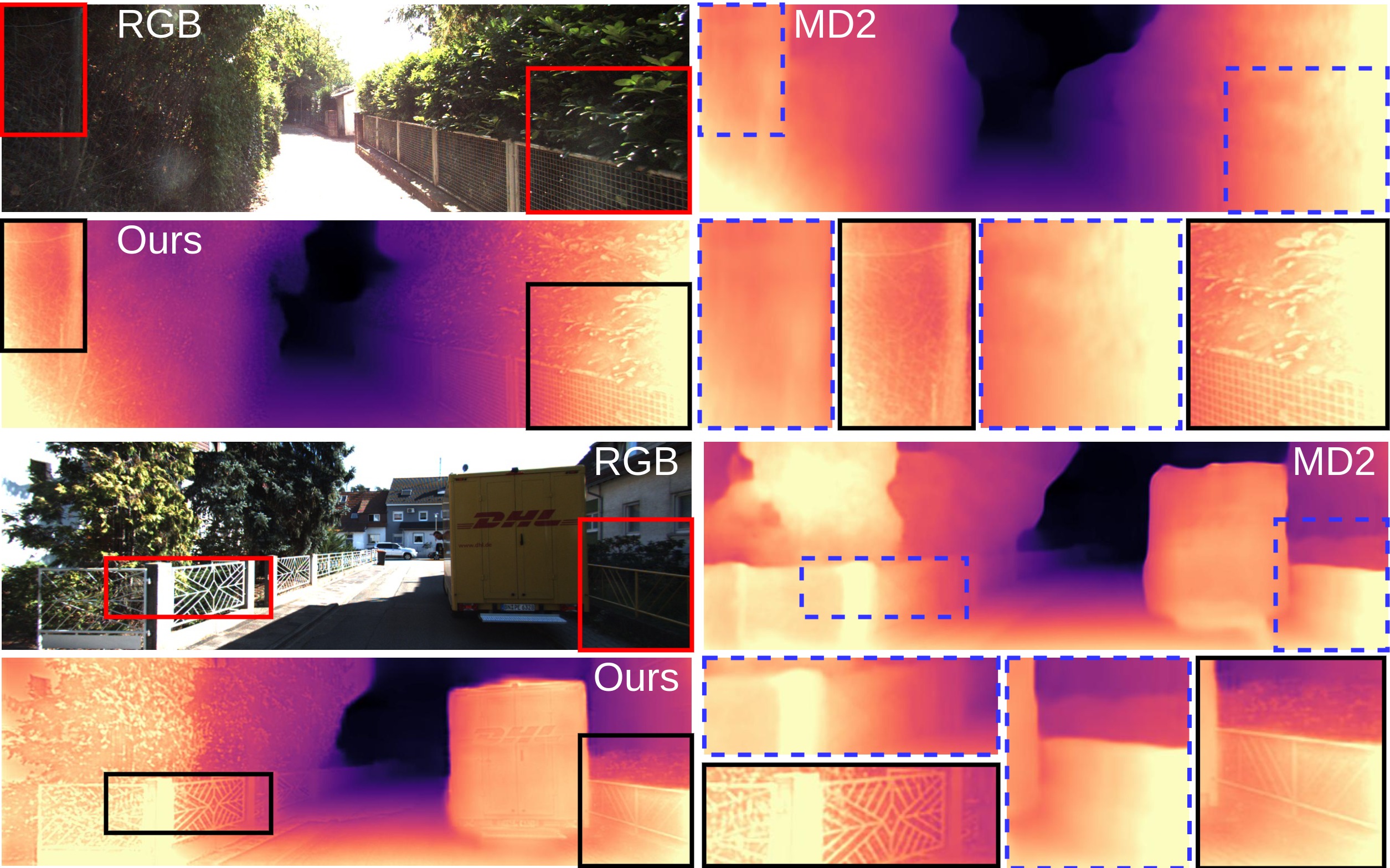}
       \caption{Here, we show clear examples where BaseBoostDepth (black solid border) outperforms Monodepth2 (blue dashed border) in capturing sharp details around fine regions. Thin mesh railings are visible with our depth network, and leaf structures emerge more clearly.}
   \label{fig:vis}
\end{figure*}

Finally, we present qualitative results in Figure \ref{fig:vis}, which illustrate the advantages of our depth network over prior methods, owing to our superior depth edge accuracy.
\section{Conclusion}
Our study demonstrates leveraging wider baselines for improved self-supervised monocular depth estimation, enhancing image, edge, and point cloud metrics. Traditionally, wider baselines were avoided in depth estimation due to perceived limitations and the oversight of wider baselines when using minimum aggregation. However, by implementing our curriculum-learning-inspired strategy, and carefully guiding the pose estimations, we extract great benefits for edge-based depth improvements. Our boosting strategy is depth backbone-agnostic and can be initialized from the warm-up phase or with pre-trained weights. Additionally, our improvements do not result in any increase in computational cost at test time. We anticipate that our findings will advance research toward more refined detail adaptation in the future.
\section{Acknowledgement}
This research was funded and supported by the EPSRC's DTP, Grant EP/W524566/1. Most experiments were run on Aston EPS Machine Learning Server, funded by the EPSRC Core Equipment Fund, Grant EP/V036106/1.

\bibliography{main}

\clearpage
\begin{center}
    \Large\textbf{\textcolor[rgb]{0,0,.4}{Supplementary Materials}}
\end{center}
\section{Hyperparamter Tuning}
\label{sec:hyper}
\begin{table*}[ht]
  \centering
  \tiny
  \resizebox{1\textwidth}{!}{
    \begin{tabular}{|l||c|c|c|c|c|c|c|c|c|}
        \arrayrulecolor{black}\hline
           & \multicolumn{7}{c|}{KITTI} & \multicolumn{2}{c|}{SYNS} \\

          $\alpha$ & \cellcolor{col1}Abs Rel & \cellcolor{col1}Sq Rel & \cellcolor{col1}RMSE  & \cellcolor{col1}RMSE log & \cellcolor{col2}$\delta < 1.25 $ & \cellcolor{col2}$\delta < 1.25^{2}$ & \cellcolor{col2}$\delta < 1.25^{3}$ & 
        \cellcolor{col1}Acc & 
        \cellcolor{col1}Comp \\
         
        \hline\hline

         5.0 &   \underline{0.107}  &   0.738  &   \textbf{4.584}  &   \textbf{0.184} &   \textbf{0.883}  &   \textbf{0.963}  &   \textbf{0.983}  & \textbf{2.449}  &   \textbf{3.734}  \\
        \hline
        
         5.25 &     \underline{0.107}  &   \textbf{0.733}  &   4.592  &   \textbf{0.184}  &   \textbf{0.883}  &   \underline{0.962}  &   \textbf{0.983} &    \underline{2.452}  &   \underline{3.804}   \\
        \hline
        
        5.5 &    \textbf{0.106}  &   \underline{0.736}  &   \textbf{4.584}  &   \textbf{0.184}  &   \textbf{0.883}  &   \textbf{0.963}  &   \textbf{0.983} &   {2.453}  &   {3.810} \\ 
        \hline
        
        5.75   &   \underline{0.107}  &   0.739  &   \underline{4.587}  &   \textbf{0.184}  &   \textbf{0.883}  &   \underline{0.962}  &   \textbf{0.983}   &   2.459  &   4.177 \\ \hline
         
        6.0   &   \underline{0.107}  &   0.747  &   \underline{4.587}  &   \textbf{0.184}  &   \textbf{0.883}  &   \textbf{0.963}  &   \textbf{0.983}   &   2.455  &   3.889 \\
        \hline
        
        \arrayrulecolor{black}\hline

    \end{tabular}}  
  \caption{\textbf{Hyperparameter tuning $\alpha$.} We conducted depth evaluations on the KITTI and SYNS dataset using the same parameters as described in the main paper. This extends from the results presented in row 7 of Table 1 in the main paper, where we incorporated error-induced reconstruction into our final model.}

  \label{tab:hypers}
\end{table*} 
In the ablation study, we introduce error-induced reconstruction, which introduces a controlled error in the pose estimation for image reconstruction. In Table \ref{tab:hypers}, we present the tuning of $\alpha$ for selecting the pose error. Our objective is to achieve a balance between image-based metrics for the KITTI dataset and edge-based metrics from the SYNS dataset.

Table \ref{tab:hypers} shows how setting the pose error parameter $\alpha$ to 5.5 yields the most favorable image-based metrics. However, when using $\alpha=5$, the best edge detail was observed. We selected $\alpha=5.5$ to enable \textbf{BaseBoostDepth} to surpass the baseline in image-based metrics and achieve improved results in edge-based metrics compared to row 7 in Table 1 from the main text. However, this parameter can be fine-tuned to emphasize edge-based metrics more prominently, depending on the application.
\section{Incremental Pose Estimation}
\begin{figure}[ht]
  \centering
   \includegraphics[width=1\linewidth]{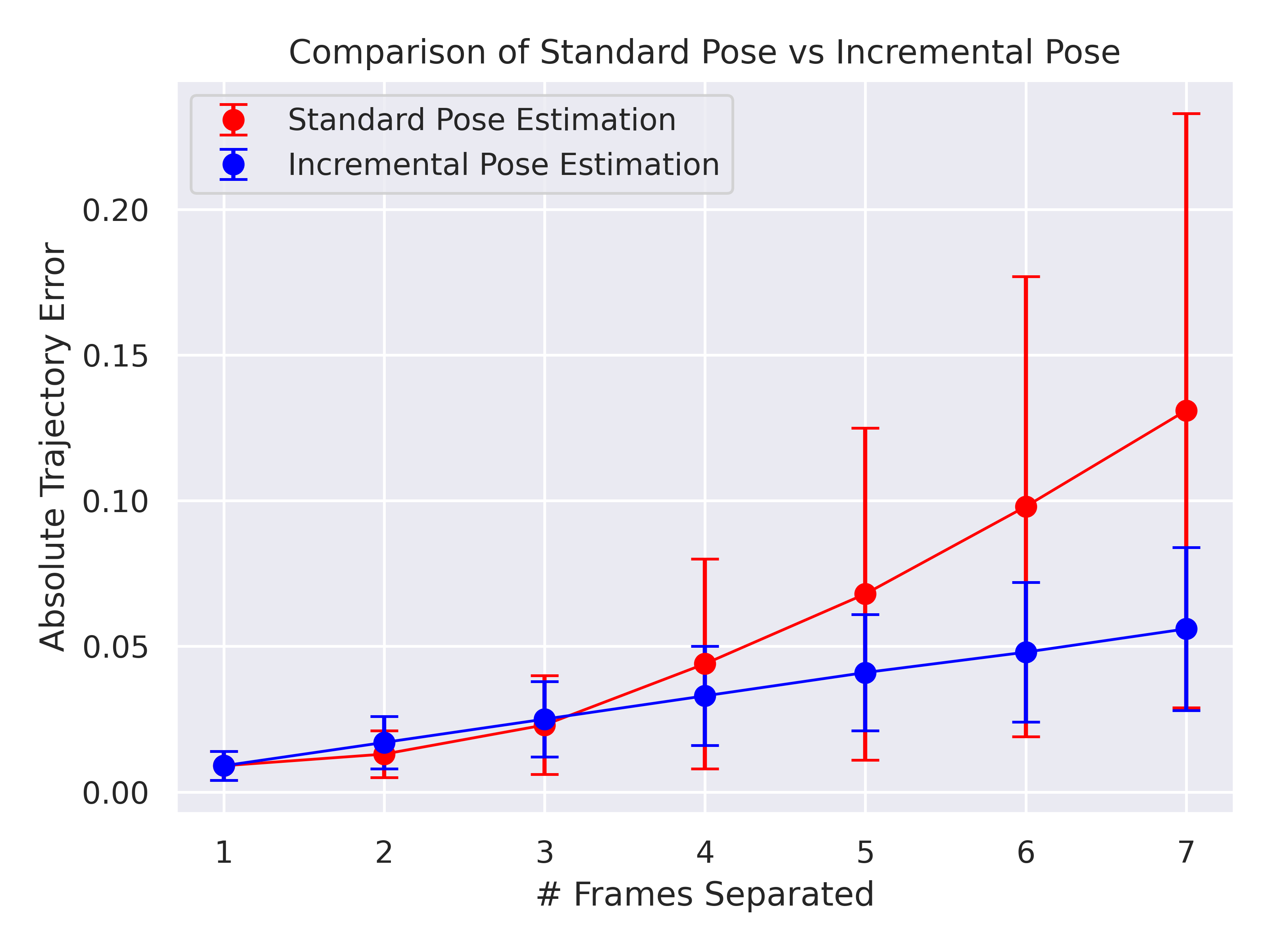}
   \caption{In red, we depict Monodepth2's pose error when calculating pose over \# Frames Separation. In blue, we illustrate the error of the pose when employing incremental pose estimation. Incremental pose estimation exhibits linear increases in error, while the standard method show exponential increases in error for larger frames of separation. In this analysis, we test using ground truth pose on sequence 9 of the KITTI odometry split and average over one frame.}
   \label{incremental}
    \vspace{-0.4cm}
\end{figure}

We now provide justification for our use of incremental pose estimation, as discussed in Section 3.4 of the main paper. We observed that pose estimation drifts occur with larger frame separations, leading to incorrect pose estimations and increased average absolute trajectory error. In Figure \ref{incremental}, we present the baseline (Monodepth2) performance when using standard pose estimation as the number of frames of separation between the source and the target images vary. Additionally, we show the impact of incremental pose estimation on reducing the error of pose estimation over larger frame separations.
The figure indicates that the error in pose increases exponentially after three frames of separation for the standard method of pose estimation. However, with incremental pose estimation, we observe a linear increase in error and much smaller variance in the mean error. This validates our assertion of improved pose error when using incremental pose estimation. However, the error for larger frame separations still increases using this method. We believe that further refinement of the pose estimation could mitigate errors at larger frames of separation, a topic we plan to explore in future work.
\section{Ablation Visualisation}
\label{sec:abs}
\begin{figure}[ht]
  \centering
   \includegraphics[width=1\linewidth]{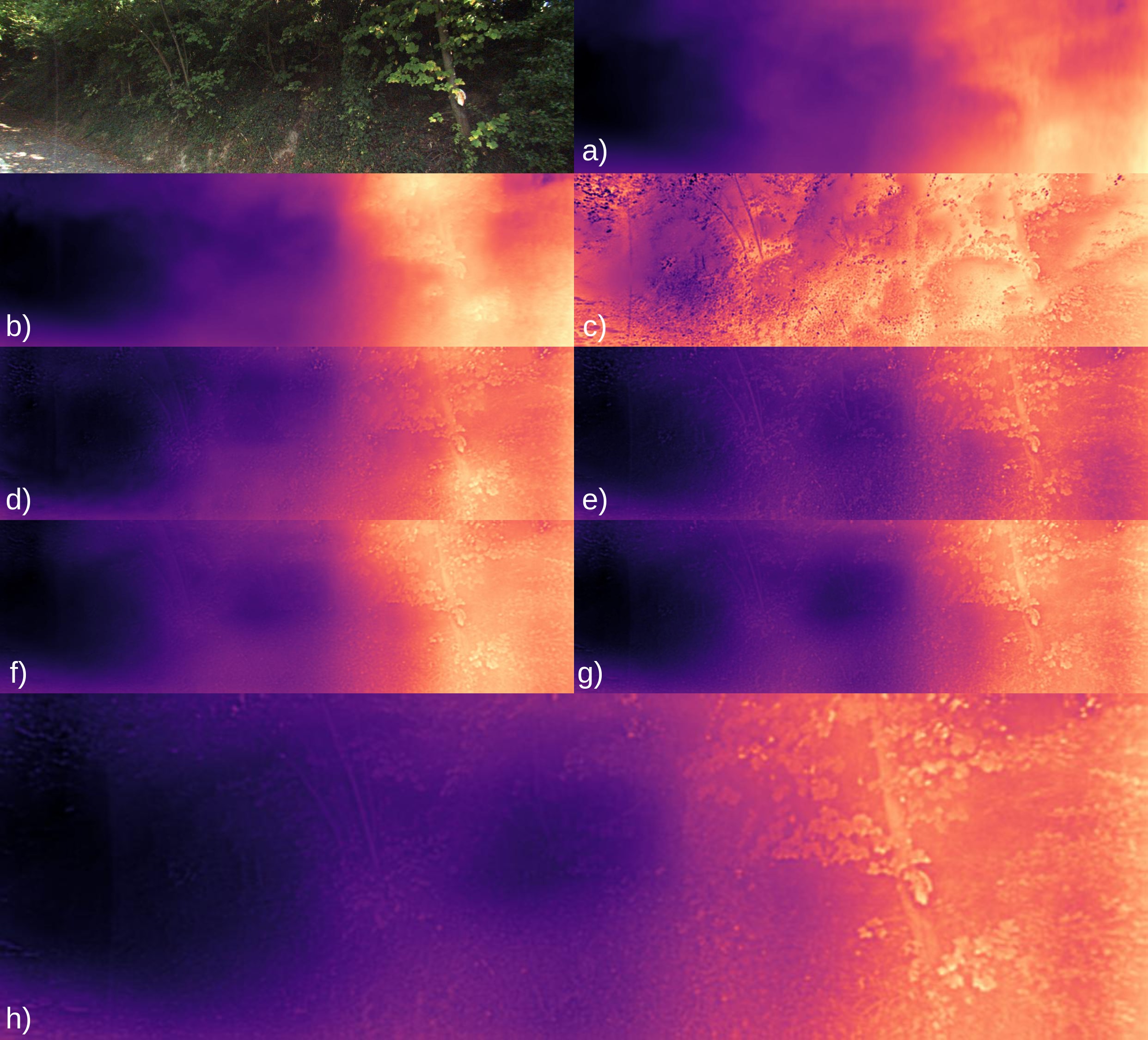}
   \caption{Contributions from the ablation study. We show, rows 1 through 8 from the the ablation study in order of alphabet. Where (a) starts with Monodepth2 as the baseline and (h) represents all contributions.}
   \label{fig:ablation_vis_nice}
    \vspace{-0.4cm}
\end{figure}

In Figure \ref{fig:ablation_vis_nice}, we show the qualitative effect of each contribution on the baseline Monodepth2. We represent the contributions as in the ablation study, which adds visual support to our edge-based metrics. 
In the ablation study, employing larger baselines, as seen in row 3 and image (c), enhances the boundaries in depth estimation. However, these improved edge-based metrics reveal noticeable areas of high brightness-contrast that are close appear further away. Furthermore, our visual inspection of rows 5 to 6 (images (e) to (f)), indicates a reduction in visual boundaries, supporting our observation that edge-based metrics decrease when employing incremental pose estimation. We regain this performance in (g) with partial incremental pose and further accentuate the edges with error-induced reconstructions in (h). 
\section{Evaluation Metrics}
\label{sec:metrics}
Within this section, we present the metrics used in the tables to evaluate depth with each dataset. We define the ground truth depth as $d$ and the predicted depth as $\hat{d}$ for both datasets.
\subsection{KITTI Eigen}
For the KITTI test set \cite{eigen2015predicting}, the metrics are as follows:

\noindent
\textbf{Absolute Relative Error (AbsRel):} This measures the mean relative error (\%) defined by:
\begin{equation}
e=\frac{1}{N}\sum_{i=1}^N \frac{|\hat{d}_i - d_i|}{d_i}.
\end{equation}

\noindent
\textbf{Squared Relative Error (Sq Rel):} This calculates the mean relative square error (\%) given by:
\begin{equation}
e=\frac{1}{N}\sum_{i=1}^N \frac{|\hat{d}_i - d_i|^2}{d_i}.
\end{equation}

\noindent
\textbf{Root Mean Square Error (RMSE):} It measures the root mean square error (meters) represented as:
\begin{equation}
e=\sqrt{\frac{1}{N}\sum_{i=1}^N |\hat{d}_i - d_i|^2}.
\end{equation}

\noindent
\textbf{Root Mean Square Log Error (RMSE Log):} This computes the root mean square log error (log-meters) as:
\begin{equation}
e=\sqrt{\frac{1}{N}\sum_{i=1}^N |\log{(\hat{d}_i)} - \log{(d_i)}|^2}.
\end{equation}

\noindent
\textbf{Threshold Accuracy:} It assesses the threshold accuracy (\%) for $\delta<1.25^k$, where $k \in {1,2,3}$, as:
\begin{equation}
\delta < 1.25^k = \frac{1}{N}\sum_{i=1}^N \Big(\text{max}\Big(\frac{\hat{d}_i}{d_i}, \frac{d_i}{\hat{d}_i}\Big) < 1.25^k\Big).
\end{equation}
\subsection{SYNS-Patches}
The SYNS dataset does not have easy to use ground truth data. In respect of reproducibility, we will release the exact steps used to create the ground truth data and the evaluation code on our GitHub page.
For the SYNS-Patches dataset \cite{adams2016southampton}, we employ 3D metrics to evaluate how well our depth maps reconstruct the 3D world. Following the methodology of \cite{ornek20222d} and its specific application by \cite{spencer2022deconstructing}, we utilize the following metrics:

\noindent
\textbf{Chamfer Distance:} This computes the Chamfer distance between the reconstructed point clouds, measured in meters, as:
\begin{equation}
e = \frac{1}{|Q|}\sum_{ \textbf{q} \in Q} \min_{\hat{\textbf{q}} \in \hat{Q}} || \textbf{q} - \hat{\textbf{q}} || + \frac{1}{|\hat{Q}|}\sum_{ \hat{\textbf{q}} \in \hat{Q}} \min_{ \textbf{q} \in Q} || \textbf{q} - \hat{\textbf{q}} ||.
\end{equation}
Where $Q$ and $\hat{Q}$ denote the ground truth and predicted point clouds, respectively, and $\textbf{q}$ and $\hat{\textbf{q}}$ represent the 3D points in the point clouds.

\noindent
\textbf{Precision:} This quantifies the percentage of predicted points within a distance threshold $\delta=0.1$ to the ground truth surface as:
\begin{equation}
P = \frac{1}{|\hat{Q}|}\sum_{ \hat{\textbf{q}} \in \hat{Q}} \Big[ \min_{ \textbf{q} \in Q} || \textbf{q} - \hat{\textbf{q}} || < \delta \Big].
\end{equation}

\noindent
\textbf{Recall:} This quantifies the percentage of ground truth points within a distance threshold $\delta=0.1$ to the predicted surface as:
\begin{equation}
R = \frac{1}{|Q|}\sum_{ \textbf{q} \in Q} \Big[ \min_{\hat{\textbf{q}} \in \hat{Q}} || \textbf{q} - \hat{\textbf{q}} || < \delta \Big].
\end{equation}

\noindent
\textbf{F-Score:} This computes the harmonic mean of precision and recall (\%) as:
\begin{equation}
F = 2 \cdot \frac{P \cdot R}{P+R}.
\end{equation}

\noindent
\textbf{Intersection over Union (IoU):} It quantifies the volumetric quality of a 3D reconstruction (\%) as:
\begin{equation}
IoU = \frac{P \cdot R}{P+R - P \cdot R}.
\end{equation}

Additionally, edge-based metrics from \cite{koch2018evaluation} are considered following \cite{spencer2022deconstructing}:

\noindent
\textbf{Accuracy (Acc):} This quantifies the accuracy of the predicted depth edges (pixels \textbf{p}) by assessing the distance from each predicted edge to the nearest ground truth edge, expressed as:
\begin{equation}
e =  \frac{1}{|P|} \sum_{P} \operatorname{EDT} \Big(\hat{\textbf{D}}_{bin}(\textbf{p})\Big):P=\{\textbf{p}|\textbf{D}_{bin}(\textbf{p})=1\}.
\end{equation}
In this equation, $\operatorname{EDT}$ denotes the Euclidean Distance Transform, with a maximum distance threshold set to $\delta=10$, while $\textbf{D}_{bin}$ denotes the binary map of depth boundaries for the ground truth, and $\hat{\textbf{D}}_{bin}$ represents the binary map of predicted depth boundaries.

\noindent
\textbf{Completeness (Comp):} It quantifies the completeness of the predicted depth edges (pixels \textbf{p}) as the distance from each ground truth edge to the nearest predicted edge, expressed as:
\begin{equation}
e = \frac{1}{|P|} \sum_{P} \operatorname{EDT} \Big({\textbf{D}}_{bin}(\textbf{p})\Big):P=\{\textbf{p}|\hat{\textbf{D}}_{bin}(\textbf{p})=1\}.
\end{equation}
All the metrics outlined above were informed by the study conducted by \cite{spencer2022deconstructing}, which extensively analyzed the SYNS dataset.
\section{Eigen Benchmark}
\label{sec:bench}
We also present performance on the improved ground truth KITTI data in Table \ref{tab:kitti_eigen_bench}. We remain competitive in all metric against previous methods when using the equivalent backbones. It is important to note that variants of \textbf{BaseBoostDepth} generally improve image-based depth estimation while also exhibiting enhanced edge-based and point-cloud-based results with the SYNS datasets.
\begin{table*}[ht]
  \centering
  \tiny
  \resizebox{1.0\textwidth}{!}{
    \begin{tabular}{|l||c|c|c|c|c|c|c|}
        \arrayrulecolor{black}\hline
          Method & \cellcolor{col1}Abs Rel & \cellcolor{col1}Sq Rel & \cellcolor{col1}RMSE  & \cellcolor{col1}RMSE log & \cellcolor{col2}$\delta < 1.25 $ & \cellcolor{col2}$\delta < 1.25^{2}$ & \cellcolor{col2}$\delta < 1.25^{3}$ \\
         
        \hline\hline
        
        Monodepth2~\cite{godard2019digging}  &   {0.075}  &   0.410  &   {3.495}  &    {0.119}  &   {0.934}  &   {0.987}  & {0.996} \\ \hline

        CADepth~\cite{yan2021channel} &   0.072  &   0.381  &   3.352  &   0.113  &   {0.940}  &   0.989  &   \underline{0.997}  \\
        \hline
              
        DIFFNet ~\cite{zhou2021self}&   {0.071}  &   0.360  &   {3.316}  &   \underline{0.111}  &   {0.943}  &   {0.990}  &   \underline{0.997}  \\
        \hline
                 
        MonoViT~\cite{zhao2022monovit}   &   \textbf{0.066}  &   \underline{0.320}  &   \underline{3.136}  &    \textbf{0.104}  &   \underline{0.950}  &   \underline{0.991}  &   \textbf{0.998}  \\
        
        \hline\hline
                
        \textbf{BaseBoostDepth}     &   0.078  &   {0.407}  &   3.613  &   0.124  &   0.927  &   {0.986}  &   {0.996}  \\
        \hline
        \textbf{BaseBoostDepth$_{pre}$}     &   0.076  &   0.393  &   3.519  &   0.120  &   0.932  &   0.987  &   \underline{0.997}  \\
        \hline

        \textbf{BaseBoostDepth$_{pre}^\dagger$}     &   \underline{0.068}  &   \textbf{0.307}  &   \textbf{3.097}  &   \textbf{0.104}  &   \textbf{0.951}  &   \textbf{0.992}  &   \textbf{0.998}  \\

        \hline
        
        
        \arrayrulecolor{black}\hline

    \end{tabular}}  
  \caption{\textbf{Quantitative Results for the KITTI Eigen Improved Test Dataset.} Note that we do not use any test time refinement and only require one frame to do inference for depth estimation. \textbf{Bold} text represents the best result for the metric in each group.}

  \label{tab:kitti_eigen_bench}
\end{table*}
\section{Ethical Considerations}
In this section we discuss the ethical considerations of using both the KITTI \cite{geiger2012we} and SYNS-Patches\cite{adams2016southampton} datasets. Individuals represented in these datasets, such as pedestrians, cyclists, and drivers, may not have explicitly consented to the collection and use of their data for research purposes. As a result, the dataset presents privacy concerns due to the unintentional inclusion of individuals, potentially infringing upon their privacy rights. Furthermore, the presence of registration plates exacerbates these privacy issues. Moreover, since the data is primarily sourced from Western regions, it becomes challenging to extrapolate findings to other geographical locations and cultural contexts.
\section{Brightness-Contrast Cues \& Limitations}
In this section, we aim to substantiate our claim that our depth network exhibits stronger brightness-contrast cues compared to baseline methods, resulting in sharper object boundaries. Additionally, we will address the limitations associated with relying heavily on this cue.

\begin{figure}[h!]
  \centering
   \includegraphics[width=1\linewidth]{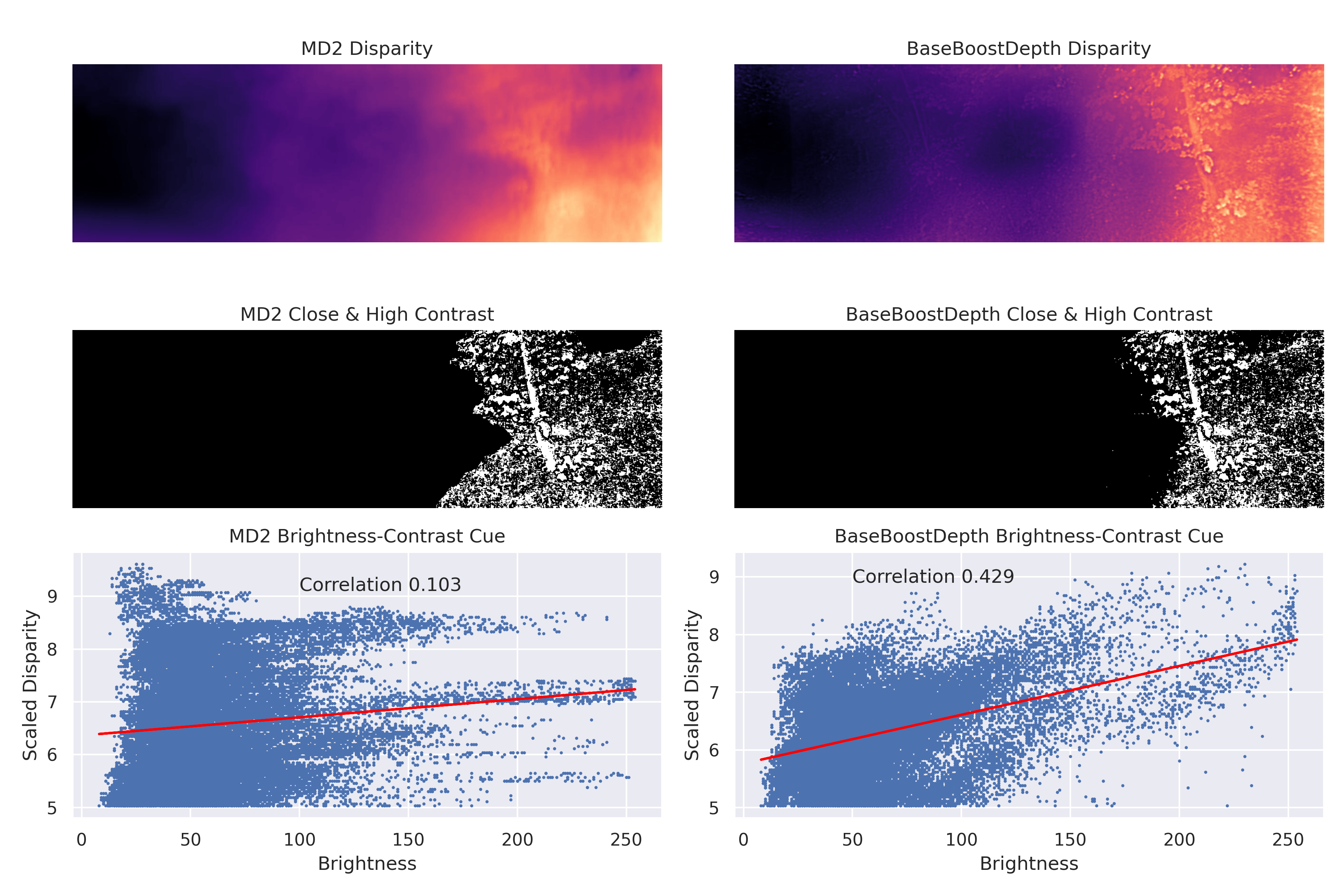}
   \caption{We find that \textbf{BaseBoostDepth}, in high contrast and closer regions, has a stronger correlation between brightness and scaled disparity than Monodepth2. }
   \label{fig:limit_quan}
\end{figure}

We primarily observe this effect in foreground objects, specifically those relatively close to the camera. As depicted in Figure \ref{fig:limit_quan}, we demonstrate a high correlation between brightness and scaled disparity values in high contrast regions, particularly those near the camera. In the provided image, we consider \textit{``close''} to be half of the maximum observed distance from the LiDAR, although this will vary for each image.
The first row illustrates the respective disparity maps, while the second row displays binary masks representing high contrast regions within close proximity. The scatter plots in the final row depict the relationship between brightness value and scaled disparity value. Notably, \textbf{BaseBoostDepth} exhibits a moderate correlation between brightness and disparity values, suggesting that the brightness-contrast cue is four times more impactful than that for Monodepth2. It is notable that we observe this pattern occurring primarily in high-frequency regions, with little to no noticeable effect in low-frequency regions.

\begin{figure}[ht]
  \centering
   \includegraphics[width=1\linewidth]{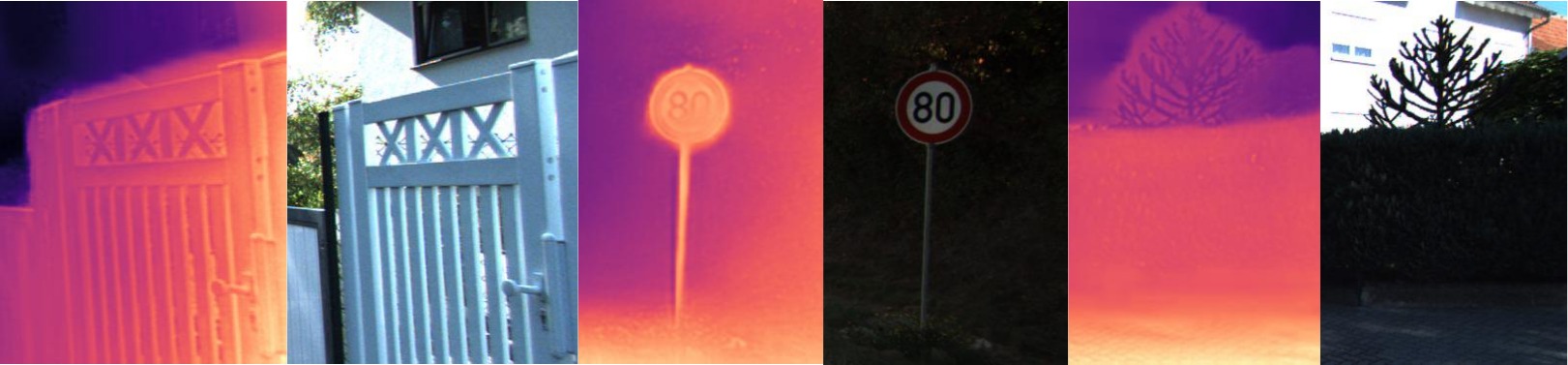}
   \caption{While the brightness-contrast cue proves beneficial in certain scenes, there are some regions where these cues mislead the depth network. }
   \label{fig:limits}
    \vspace{-0.4cm}
\end{figure}

As our model exploits brightness-contrast cues, it can overemphasize high contrast regions, providing closer estimates for bright pixels. This trend is notably evident in the example depicted on the far right of Figure \ref{fig:limits}, where tree branches, despite being foreground objects, are perceived as more distant than the white background due to their contrast.

Similar issues arise with road signs and lane markings, potentially posing significant hazards in driving scenarios, as lane lines could be misinterpreted as road irregularities. Despite our attempts to mitigate this issue using inverse augmentations, we were unable to find a satisfactory solution. Consequently, we leave this problem for future research endeavors. As a final note, although we acknowledge these limitations, we observe improvements in all evaluation metrics. Therefore, our model outperforms the baseline methods overall.
\section{Final Results}
\begin{figure}[ht]
  \centering
   \includegraphics[width=1\linewidth]{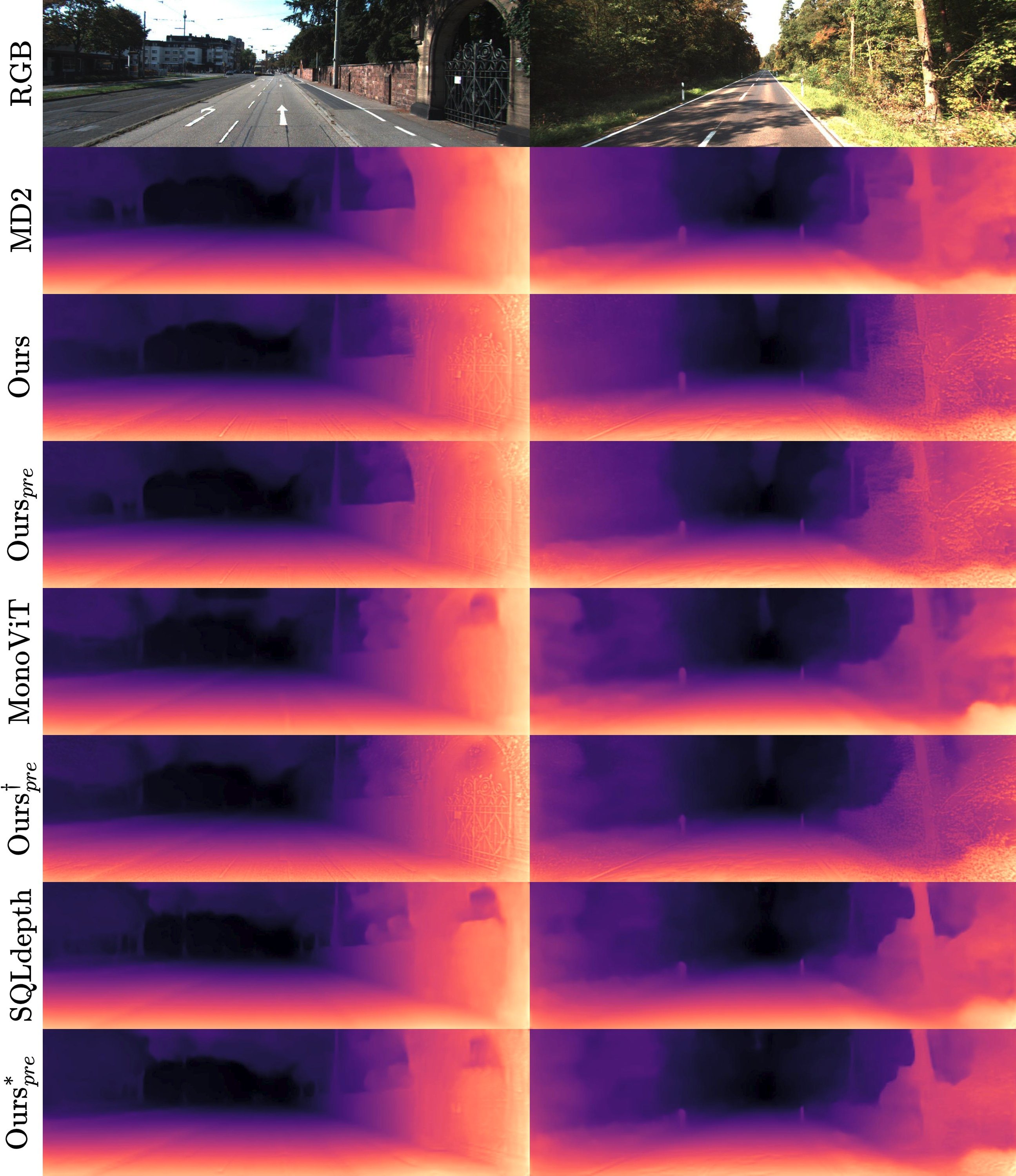}
   \caption{We show qualitative results for our model \textbf{BaseBoostDepth} (\textbf{Ours}) comparing to the baseline (MD2). We also show our pretrained version on Monodepth2 (\textbf{Ours}$_{pre}$) and MonoViT (\textbf{Ours}$_{pre}^\dagger$). All results are from the KITTI test split.}
   \label{fig:final1}
\end{figure}

\begin{figure}[ht]
  \centering
   \includegraphics[width=1\linewidth]{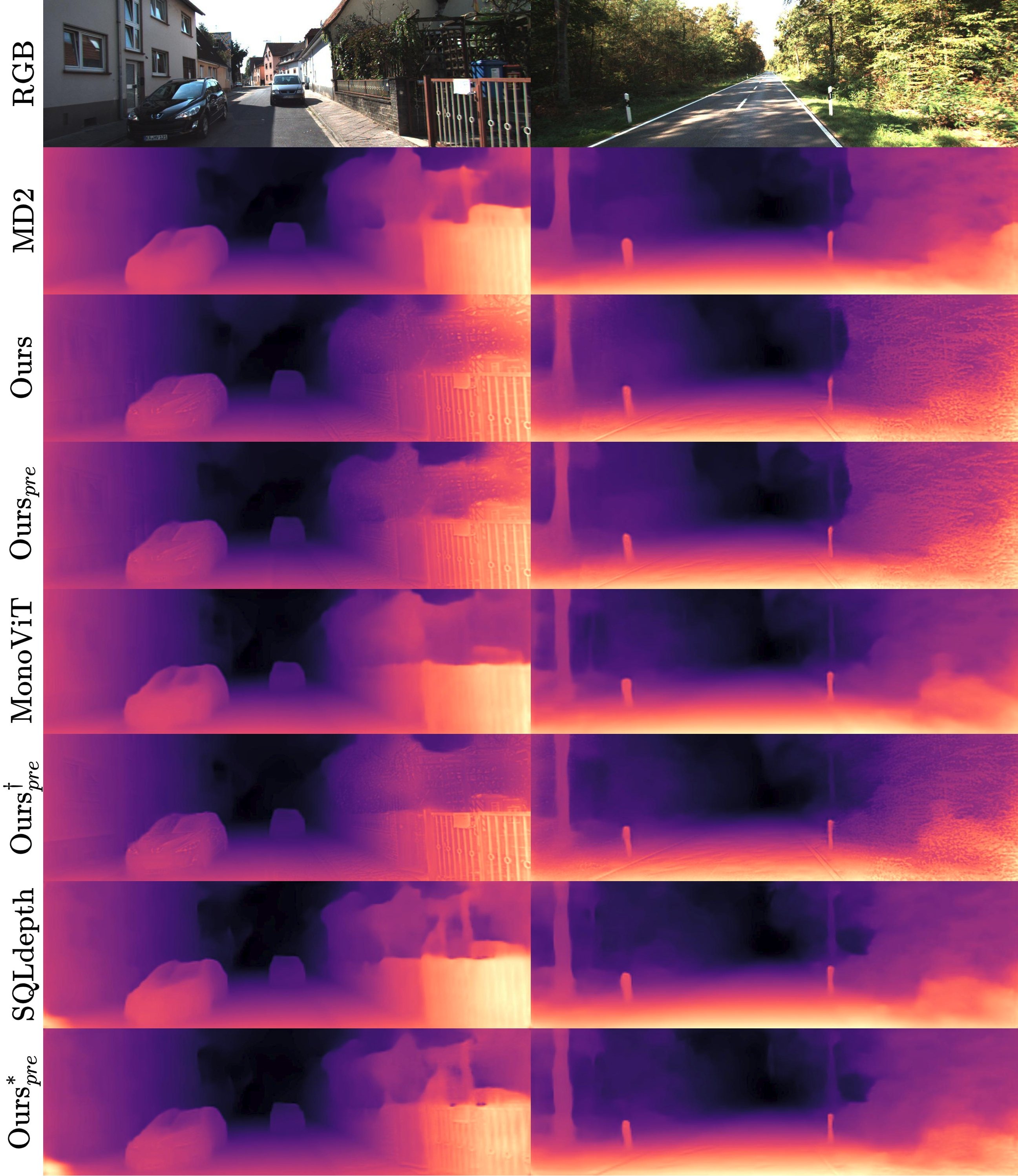}
   \caption{We show additional qualitative results for our models showing sharper definition on fence structures and shrubbery.}
   \label{fig:final2}
\end{figure}

Finally, we present the qualitative results showing the benefits of our models in Figure \ref{fig:final1} and Figure \ref{fig:final2}. Namely we show \textbf{BaseBoostDepth} (\textbf{Ours}), \textbf{BaseBoostDepth$_{pre}$}~(\textbf{Ours$_{pre}$}) and \textbf{BaseBoostDepth$_{pre}^\dagger$}~(\textbf{Ours$_{pre}^\dagger$}), in comparison to Monodepth2 \cite{godard2019digging} and MonoViT \cite{zhao2022monovit}. Our depth networks produces sharper edge definitions and reveals finer details that are often blurred in current State-of-the-Art methods.

\end{document}